\documentclass[9pt,a4paper,twocolumn,twoside]{rho-class/rho}
\usepackage[english]{babel}
\usepackage{booktabs}
\usepackage{multirow}
\usepackage{amsmath}
\usepackage{amssymb}

\doctype{Research Article}
\title{CTRAG: An In-Context Retrieval-based Framework for Automated Compliance Checking using LLMs}

\author[a, b]{Muhammad Roman} 
\author[b]{Karen Rafferty} 
\author[b]{Barry Devereux}
\affil[a]{Bristol Research and Innovation Laboratory (BRIL), Toshiba Europe Ltd., Bristol, United Kingdom}
 
\affil[b]{Queen's University Belfast, United Kingdom} 

\journalname{arXiv Preprint} \journal{arXiv Preprint} \theday{\today}

\begin{abstract}
Trust is fundamental in modern regulatory ecosystems, and compliance checking plays a critical role in fostering that trust. Regulatory compliance verification is essential for businesses operating in highly controlled environments, as it ensures alignment with sector-specific guidelines across domains such as financial reporting, data privacy, and cybersecurity. Manual compliance testing, however, is often time-intensive and prone to inconsistencies, particularly when compliance depends indirectly on third-party services such as cloud providers, where vendors rely on external providers to meet regulatory standards. In this paper, we present CTRAG, a novel Retrieval-Augmented Generation (RAG) pipeline designed for automated compliance checking. CTRAG employs advanced strategies, including adaptive chunking, dynamic retrieval configurations, and in-context learning, to improve the precision and relevance of compliance assessments. By extracting control questions from regulatory texts and cross-referencing them with unstructured company documentation, CTRAG achieves highly accurate, document-informed compliance verification, even in cases of indirect compliance through third-party services. Empirical evaluations demonstrate significant improvements, with CTRAG achieving an F1-score of 78\% and a recall of 85\% in the final deployed configuration, ensuring minimal missed non-compliance cases while reducing manual reviewer effort in a real-world deployment. To validate CTRAG value, we developed and deployed a POC within a Big Four professional services firm, applying it to real-world cases and cross-checking results against manual compliance reports. These findings highlight CTRAG potential to streamline compliance workflows, mitigate risks, and enhance regulatory trust in complex, high-stakes environments.
\end{abstract}


\begin{document}
    \maketitle
    \thispagestyle{firststyle}

\section{Introduction}
\label{section:into}
In the increasingly regulated corporate landscape, maintaining compliance with sector-specific controls has become essential yet challenging for organizations. Regulatory bodies require companies to adhere to established controls, which are often complex and multifaceted, spanning across various domains such as financial reporting, data privacy, environmental standards, and cybersecurity. Compliance with these regulatory controls is typically validated through extensive document audits, requiring organisations to present evidence that their operations align with regulatory expectations. In such high-stakes environments, effective and efficient compliance testing is essential to minimise risks and ensure adherence to the regulations.

\begin{figure*}
  \centering
  \includegraphics[width=1\linewidth]{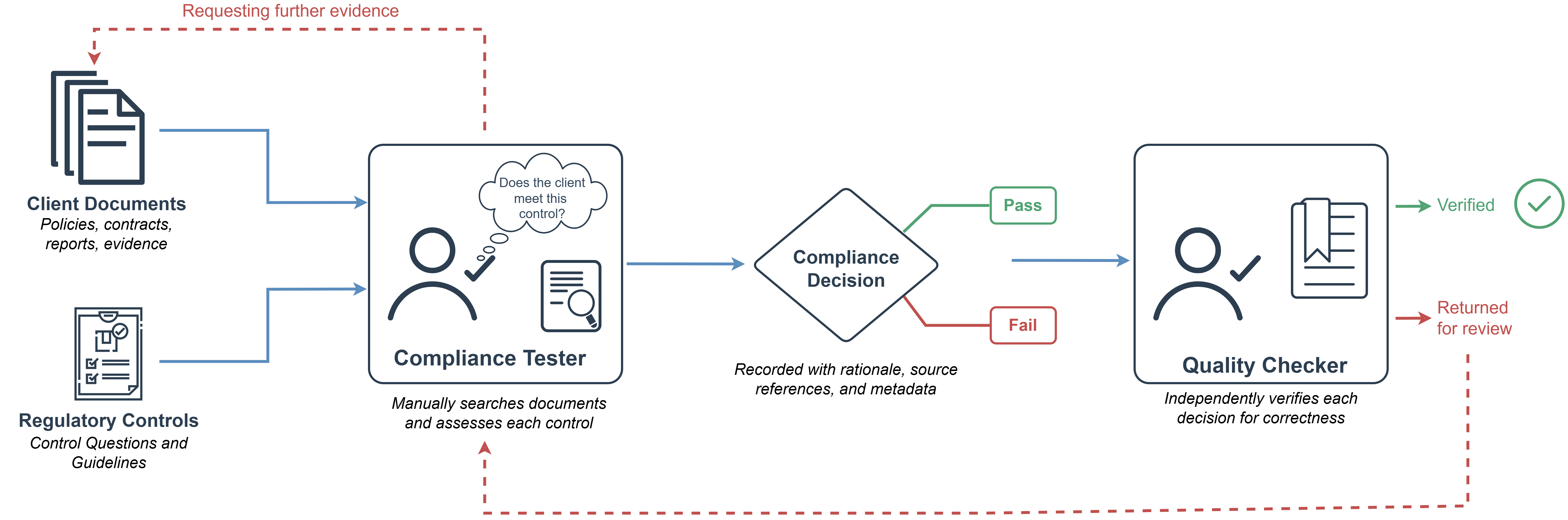}
  \caption{Flow of third party compliance checking performed manually by a compliance tester and validated}
  \label{fig:manualCT}
\end{figure*}

Traditional compliance testing often involves labour-intensive manual reviews, depicted in Figure \ref{fig:manualCT}, where auditors analyse substantial volumes of documentation provided by companies. This process, while thorough, is time-consuming and can introduce variability in results due to differing interpretations of regulatory language. Compliance reviews require auditors to locate relevant information across extensive collections of company documents. New materials are added with each case, which substantially increases both the complexity and the risk of the review. With advancements in Natural Language Processing (NLP) and the emergence of Retrieval-Augmented Generation (RAG) \cite{lewis2020retrieval} models, there is a promising opportunity to streamline compliance testing by leveraging machine learning (ML) to automate the retrieval and evaluation of relevant compliance evidence.

We introduced a RAG-based pipeline tailored specifically for regulatory compliance testing. Using this structured representation, our approach systematically verifies a company's compliance by analysing and cross-referencing these controls with documents provided for audit. The pipeline incorporates several layers of customisation to optimise response accuracy and relevance. RAG helps both in retrieving the right information from the documents, and in using that information for regulatory compliance tests. We begin by extracting and processing information from company-provided PDF documents, applying a variety of chunking strategies that maintain logical coherence while minimizing extraneous information. We dynamically adjust chunk sizes based on the type of document, leveraging methods that ensure both contextual integrity and retrieval accuracy. We also adjust the k value retrieving the top-k relevant documents that can potentially be used to look for the compliance check. This helps provide access to the right information that plays a vital role in identifying compliance, as witnessed by experiments. Finally, we integrate in-context learning (ICL) \cite{brown2020language, zhou-etal-2024-mystery}, enriching the prompt with curated examples relevant to the queries. This significantly improved response relevance and alignment with regulatory expectations.
This paper evaluates the efficacy of this RAG-based compliance testing pipeline and its contributions to reducing audit time while maintaining compliance accuracy. Our findings aim to demonstrate how automated compliance verification can streamline regulatory assessments and mitigate risks associated with non-compliance in complex regulatory environments.

The main contributions of this paper are as follows:

\begin{enumerate}
    \item The paper introduces a RAG pipeline specifically tailored for regulatory compliance verification. This pipeline integrates the extraction of \textit{controls} from regulatory documents to enable precise, document-informed checks against client-provided files containing unstructured contents.

    \item The study benchmarks chunking, and number of chunks to find the best fit by extensive experimentation that helps research community to understand the impact of chunk size variation in similar problems. We have also investigated the effects of various LLM models on the final evaluation.
    
    \item It employs several adaptive strategies to enhance retrieval accuracy and response quality, including robust PDF information extraction, tailored chunking techniques, and context extraction mechanisms.

    \item The pipeline uses a hybrid retrieval system to balance global regulatory context with granular factual details, dynamically setting retrieval K-values to ensure contextually relevant documents contribute to response generation.

    \item The approach employs in-context learning to refine responses by providing carefully selected examples that guide the LLM's decision-making process.
    
\end{enumerate}

\section{Related Work}
\label{section:related}

Automated compliance checking (ACC) has been an active area of research for several decades, driven by the growing complexity of regulations and the increasing cost of manual compliance assessment. Early work focused primarily on formal and rule-based approaches that sought to represent regulations, contracts, and organisational policies in machine-interpretable formats, enabling systematic verification of compliance requirements. In the business process domain, compliance checking has been investigated as a means of assessing whether organisational processes conform to contractual obligations, legal requirements, and internal governance policies \cite{kharbili2008business}. Subsequent research expanded these foundations by examining compliance throughout the lifecycle of business processes, including design-time verification, run-time monitoring, and post-execution auditing, while highlighting persistent challenges arising from complex regulations, evolving requirements, and the difficulty of integrating compliance mechanisms into operational systems \cite{hashmi2018}. Collectively, these studies established many of the conceptual and methodological foundations that continue to underpin contemporary compliance automation research.

Beyond business process management, automated compliance checking has been extensively investigated in the Architecture, Engineering, Construction, and Operations (AECO) sector, where regulatory verification is traditionally performed through labour-intensive manual reviews of design artefacts. The emergence of Building Information Modelling (BIM) and Industry Foundation Classes (IFC) enabled researchers to represent building information in machine-readable formats, creating opportunities for automating regulatory assessments against building codes and standards \cite{chen2024automated}. Research in this area has focused on translating textual regulations into computable rules, developing semantically rich object models, and establishing interoperability mechanisms capable of supporting regulatory reasoning throughout the building lifecycle. More recent studies have argued that effective compliance checking requires a broader digital ecosystem in which regulatory requirements, design information, and verification mechanisms interact seamlessly across multiple stakeholders and systems \cite{amor2021promise, beach2024digital}. Despite substantial progress, challenges related to the formalisation of regulatory provisions, interoperability across heterogeneous data sources, and the interpretation of complex requirements continue to limit the scalability of automated compliance solutions.

Outside the construction domain, compliance automation has become increasingly important in cybersecurity, privacy, and cloud-based service environments, where organisations must demonstrate adherence to a growing number of regulatory and industry frameworks. Recent studies have highlighted the shift from periodic auditing towards continuous compliance in cloud-based and regulated environments, where compliance-as-code and policy-as-code approaches are used to automate evidence collection, policy validation, and traceability across heterogeneous systems \cite{yanagawa2024}. In parallel, researchers have investigated the specific compliance challenges faced by Software-as-a-Service (SaaS) providers, where regulatory obligations relating to data protection, privacy, and information security must be satisfied within highly dynamic and distributed environments \cite{humayun2022software}. These studies emphasise the need for scalable and automated approaches capable of reducing the cost of compliance monitoring while maintaining consistency, traceability, and alignment with evolving regulatory requirements across diverse operational contexts.

To address the limitations of purely rule-based compliance verification, researchers increasingly explored artificial intelligence techniques for automating regulatory interpretation and compliance assessment. Early efforts employed expert systems and knowledge-based reasoning to encode domain expertise and provide automated compliance recommendations \cite{beach2015rule}. More recent work has revisited the role of artificial intelligence in compliance checking, highlighting both the opportunities and challenges of applying AI techniques to regulatory reasoning and compliance assessment \cite{jain2024, berger2023towards}. Subsequently, advances in natural language processing and machine learning enabled the extraction of normative knowledge from regulations, standards, and policy documents, reducing the dependence on manually crafted rule sets \cite{salama2016semantic, zhang2016semantic}. Ontology-based approaches further enhanced automated reasoning by providing formal semantic representations of regulatory concepts, relationships, and constraints, supporting object mapping, rule execution, and knowledge interoperability across compliance domains \cite{beach2015rule, zhou2017ontology}. More recent work has integrated these semantic foundations with machine learning, natural language processing, and AI techniques to improve regulatory interpretation and automated compliance assessment \cite{zhang2022building, alnuzha2025role}.

Despite these advances, automated compliance checking continues to face significant challenges arising from the inherently complex and often ambiguous nature of regulatory requirements. Many compliance obligations are expressed in natural language and contain implicit assumptions, contextual dependencies, and subjective terminology that are difficult to formalise computationally. Zhang et al. \cite{zhang2023unpacking} demonstrate that a substantial proportion of regulatory requirements cannot be evaluated through simple rule-based methods alone, owing to both intentional ambiguity introduced to preserve flexibility and unintentional ambiguity stemming from linguistic and domain-specific complexities. More broadly, recent studies highlight that translating complex regulatory texts into machine-interpretable representations remains a major obstacle to large-scale compliance automation, particularly when evidence must be gathered from diverse and unstructured organisational documents \cite{jain2025complexity}. These challenges have stimulated growing interest in advanced AI techniques capable of improving regulatory understanding, information extraction, and evidence-based compliance assessment.

The emergence of large language models (LLMs) has created new opportunities for addressing longstanding challenges in automated compliance checking. Unlike traditional rule-based or ontology-driven systems, LLMs can leverage advanced language understanding capabilities to interpret regulatory requirements, analyse organisational documents, and perform reasoning over complex textual evidence. Recent studies have explored the use of LLMs for regulatory interpretation, policy analysis, compliance monitoring, and automated compliance checking across multiple domains \cite{berger2023towards, jain2024, jain2025complexity, dimyadi2025leveraging}. However, concerns regarding hallucinations, evidence attribution, and explainability remain significant barriers to their adoption in high-stakes regulatory environments. To address these limitations, Retrieval-Augmented Generation (RAG) combines external information retrieval with generative language models, enabling responses to be grounded in relevant source documents rather than relying solely on parametric knowledge \cite{lewis2020retrieval}. Recent research has shown that retrieval quality, document chunking strategies, and knowledge integration mechanisms are critical determinants of RAG performance, particularly for enterprise document analysis and other knowledge-intensive tasks \cite{brown2025systematic, cheng2025knowledge}. These characteristics make RAG particularly well suited to compliance verification scenarios, where decisions must be supported by evidence extracted from large collections of organisational documentation.

Recent research has focused on improving the effectiveness of RAG systems through enhanced retrieval strategies, document chunking methods, and context integration mechanisms. Studies have shown that retrieval quality is often a primary determinant of downstream generation performance, motivating the development of specialised approaches such as Meta-Chunking, which segments text according to logical structure rather than fixed boundaries \cite{zhao2024meta}, and LongRAG, which employs hybrid retrieval mechanisms to balance broad contextual understanding with precise evidence extraction in long-document settings \cite{zhao2024longrag}. Domain-specific adaptations have also emerged, such as RAG4ITOps, which incorporates specialised retrieval and representation techniques to improve question answering over enterprise knowledge sources \cite{zhang2024rag4itops}. Similar efforts have also been reported in regulatory information retrieval and answer generation, where specialised retrieval pipelines have been developed to support evidence-grounded question answering over regulatory corpora \cite{gokhan2024rirag}. Recent studies have further demonstrated that retrieval optimisation and reranking strategies can significantly improve regulatory question-answering performance in retrieval-augmented systems operating over regulatory corpora \cite{umar2025regrag}. In parallel, research on in-context learning has demonstrated that carefully selected examples can substantially improve the reasoning capabilities of LLMs without requiring task-specific model fine-tuning \cite{brown2020language, zhou-etal-2024-mystery, huang2024raven}. Despite these advances, relatively limited research has investigated how retrieval configuration, document chunking strategies, and in-context learning can be jointly optimised for evidence-based compliance verification over unstructured organisational documents. This gap motivates the development of CTRAG, a compliance-oriented RAG framework designed to support accurate and traceable compliance assessment through document-grounded reasoning.

\section{Proposed Solution}
\label{section:proposed}
\begin{figure}[t]
\centering
\includegraphics[width=1\linewidth]{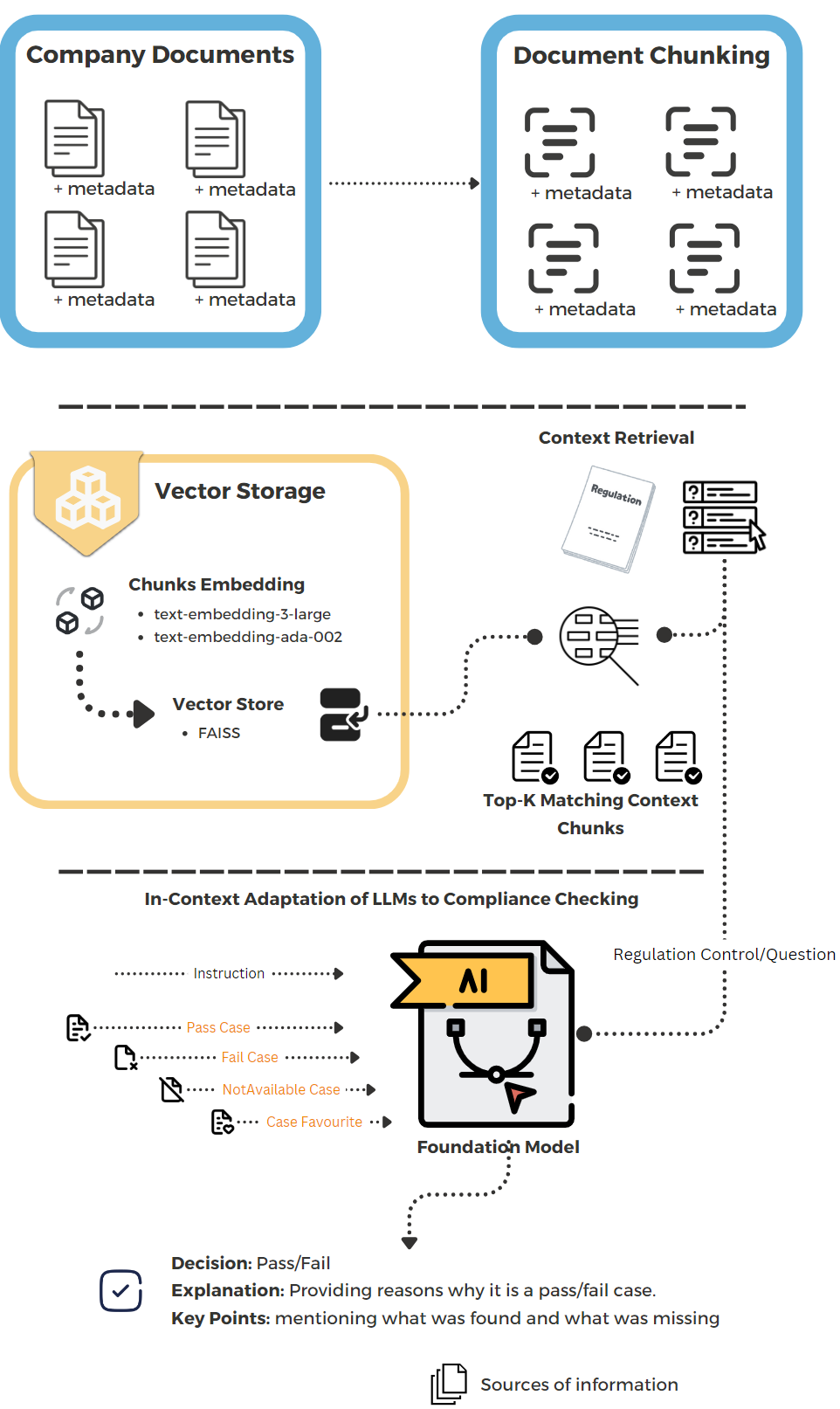}
\caption{RAG-based approach for compliance checking using clients documents and regulatory questions}
\label{fig:ragCT}
\end{figure}

Our proposed approach to automated compliance checking engages a RAG pipeline, designed to handle large, unstructured regulatory documents and company-provided files. Since compliance verification involves checking a company adherence to various regulatory controls, we frame each control as a question within a question-answering setup. This approach aligned with the RAG structure, where each control question triggers a search for the most relevant content within the company documentation, aiming to extract information that accurately addresses the compliance requirement. The multi-staged setup of RAG pipeline, as depicted in Figure \ref{fig:ragCT}, allows flexibility in configuring each component to handle the complexity and specificity of compliance testing. By combining retrieval and generation, RAG enables the system to locate precise answers in documents and present them as coherent responses, reducing the need for manual compliance audits. This approach ensures consistency and accuracy in compliance assessments while significantly reducing the time required for document review by generating responses aligned with regulatory expectations.

In our RAG-based compliance verification pipeline, document chunking is an essential process for breaking down extensive regulatory and company documents into manageable, contextually relevant segments. To achieve this, we implemented two distinct chunking strategies tailored to support precise retrieval while preserving the necessary context for compliance queries. Our first approach utilises LangChain's \textit{RecursiveCharacterTextSplitter} with adjustable window sizes and overlapping contexts, supporting both sentence-level and paragraph-level chunking. Sentence-level chunking offers fine-grained control, which is beneficial for retrieving specific clauses in response to direct compliance queries. In contrast, paragraph-level chunking captures broader contextual information, making it ideal for complex queries that rely on layered responses. This flexibility ensures that each chunk retains meaningful content without fragmenting logically connected information. In addition to LangChain text segmentation, we applied Meta-chunking \cite{zhao2024meta} to segment documents based on deeper logical relationships rather than fixed structures. Meta-chunking identifies natural boundaries within the text by grouping sentences with shared logical connections, such as causal or transitional elements. This strategy enhances retrieval accuracy by capturing more coherent segments, enabling responses that align closely with the intent and context of regulatory documents, rather than relying solely on surface-level text features.

By independently deploying these chunking methods, we ensure that retrieval operates on contextually aligned segments tailored to the diverse structures of compliance documents. This dual strategy allows us to accommodate both granular and broader context requirements, providing a foundation for accurate and relevant responses in compliance verification tasks.

To enhance compliance verification, we used query augmentation to add relevant document context to each regulatory control question. By incorporating the top-k retrieved chunks into each query, we provided a richer context that improved the alignment of responses with compliance requirements. The approach structures query inputs to allow the retrieval model to draw specific, contextually relevant information from company documents, facilitating a more accurate generation of compliance responses.

\begin{table*}[!htbp]
\caption{An illustrative example of a single control entry in the dataset, showing the control question, the ground-truth label, and the supporting documents and rationale recorded during the original manual assessment.}
\label{tab:dataset-sample}
\centering
\begin{tabular*}{\textwidth}{@{\extracolsep{\fill}}p{0.20\linewidth}p{0.72\linewidth}@{}}
\toprule
\textbf{Field} & \textbf{Example content} \\
\midrule
Control ID & C-087 \\
\addlinespace
Basic question & Do your security incident management procedures ensure that personnel are assigned roles and responsibilities for responding to incidents? \\
\addlinespace
Guidelines & Review the incident response policy and supporting procedures. Confirm that named roles (e.g., incident manager, communications lead) are defined, that responsibilities are documented, and that the policy has been reviewed within the past 12 months. \\
\addlinespace
Source documents & Information Security Policy v3.2; Incident Response Procedure; Annual Security Review Report 2023. \\
\addlinespace
Ground-truth label & Pass \\
\addlinespace
Analyst rationale (metadata only) & Roles for incident manager, technical lead, and communications lead are clearly defined in Section 4.2 of the Incident Response Procedure. Last reviewed March 2023. \\
\bottomrule
\end{tabular*}
\end{table*}

\section{Dataset}
\label{section:dataset}
The custom dataset used in this study was curated to evaluate regulatory compliance within a corporate environment. It includes 45 PDF documents that provide comprehensive insights into the company operations, covering critical areas such as security protocols, human resources policies, employee training and onboarding process, company policies, legal agreements, and annual reports. Each document is considered a potential source of information relevant to verifying the company adherence to specified regulatory standards. The dataset also comprises 240 'controls', which are essentially questions derived from regulatory guidelines. Each control has two parts: a basic question and an accompanying set of guidelines. The basic question targets specific compliance requirements, while the guidelines provide human data analysts with structured instructions on where to search for relevant information and how to assess compliance within the documents. Table \ref{tab:dataset-sample} shows an illustrative sample of the records.

For each control, the dataset also includes binary ground truth responses, labelled as 'Pass' or 'Fail,' indicating compliance or non-compliance with the corresponding regulatory standard. These responses were established by a team of expert compliance analysts, and each was further validated through quality control to ensure the reliability of the dataset answers. Additionally, each control in the original compliance assessment included open-ended commentary, detailing metadata such as the analyst responsible, the date of the compliance check, the specific documents consulted, and the rationale behind the compliance decision. However, to maintain focus and ensure clarity in evaluation, only the binary 'Pass' or 'Fail' responses were used as the ground truth answer in this study, as assessing open-ended justifications would require extensive interpretive analysis. The detailed response explanation, however, helped us in improving the response quality and guiding the model to give the right response.

RAG augments parametric memory with non-parametric retrieval and does not require model training. Therefore, the dataset is exclusively used for testing, with no portion allocated for training. This enabled a rigorous evaluation of the RAG pipeline retrieval and generation capabilities. Although smaller in scale, the dataset precise and verified responses make it a robust baseline for high-stakes compliance checking where accuracy is a high factor. The distinct structure of this dataset creates a reliable foundation for advancing compliance verification using RAG models.

\section{Experiments}
\label{section:exp}
Our experiments were designed to evaluate the impact of chunking strategies, retrieval configurations, and in-context learning techniques within our RAG-based compliance verification pipeline. The primary objective was to optimise the pipeline response accuracy, relevance, and alignment with regulatory standards. Compliance verification was framed as a question-answering task, with each regulatory control posed as a query. Through systematic experimentation, we assessed the influence of content segmentation, retrieval parameters, and response generation techniques on the pipeline overall performance.

\subsection{Chunking Strategies}
We explored multiple chunking strategies to optimise the alignment between control questions, and retrieved context. Fixed-size chunking was tested across four primary configurations: 200, 800, 1600, and 3000 characters. The 200-character configuration was excluded from later experiments due to consistently low performance; the remaining three configurations are referred to as small (800), moderate (1600), and large (3000) chars throughout the paper. The small configuration provides higher granularity, isolating specific clauses effectively for clause-level compliance queries. However, its limited context can fragment the relevant information, particularly for multi-clause compliance requirements. In contrast, the large configuration preserves broader contexts, benefiting open-ended queries but sometimes introducing irrelevant or noisy information, which impacts precision.

To address the limitations of fixed-size chunking, we implemented Meta-chunking. This approach segments documents based on logical coherence rather than fixed lengths, ensuring contextually relevant groupings. This method effectively captured topic shifts within content-rich documents, creating coherent segments that maintained logical connections across sentences. Meta-chunking excelled in aligning content with multi-layered compliance queries but incurred higher preprocessing times due to its computational complexity.

We employed OpenAI text-embedding-3-large and text-embedding-ada-002 models \footnote{https://platform.openai.com/docs/guides/embeddings}, both of which were effective in capturing compliance-specific language. Embeddings were stored in FAISS \cite{johnson2019billion}, a scalable vector store, ensuring efficient and rapid retrieval across the dataset. Persistent vector storage allowed us to minimise ingestion time when testing different generation strategies.

\subsection{Retrieval Configuration}
The number of top-retrieved documents (\(K\)) included in the query context was systematically varied to evaluate its impact on response quality. Higher \(K\) values enriched the contextual information available to the model but increased the risk of incorporating irrelevant data. Conversely, lower \(K\) values maintained precision by limiting the context but occasionally failed to retrieve key details essential for nuanced queries. Balancing \(K\) was critical for ensuring responses aligned with regulatory expectations, particularly for indirect compliance scenarios.

\subsection{Response Generation Techniques}
We experimented with different response generation methods using LangChain \textit{RetrievalQA}. Two key configurations were tested: \textit{stuff} and \textit{map\_reduce} chains. The `stuff` chain combined retrieved content into a single response, delivering efficient results for straightforward compliance checks. In contrast, the \textit{map\_reduce} chain synthesised responses from individual retrieved documents, enabling nuanced answers for complex, multi-source queries. Both methods were evaluated for their ability to produce accurate and contextually aligned responses.

\section{Results}
\label{section:results}
\begin{figure*}[h]
\centering
\includegraphics[width=1\linewidth]{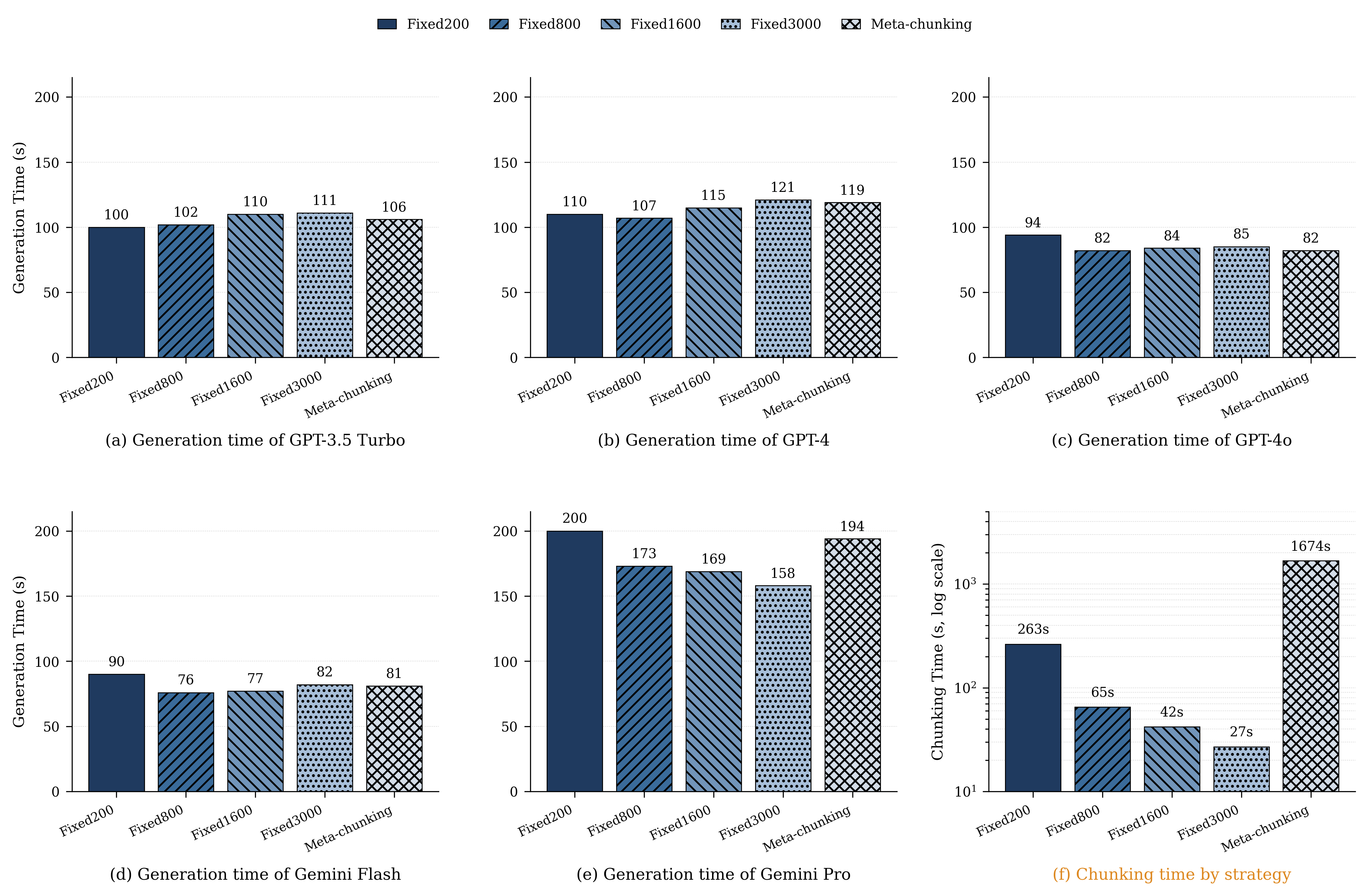}
\caption{Processing times across models and chunking strategies, with generation time (a–e) and chunking time (f).}
\label{fig:times}
\end{figure*}

The experiments conducted for compliance checking evaluated the performance of various chunking strategies, retrievals, LLMs, and in-context learning prompts, using precision, recall, and F1-score as key evaluation metrics. These experiments were designed to explore the influence of chunking configurations and model architectures on compliance-related tasks. The chunking strategies tested included fixed-size chunking with character lengths of 200, 800, 1600, and 3000, as well as meta-chunking, which groups text based on logical coherence rather than fixed size.
As shown in Figure~\ref{fig:times}, the smallest 200-character chunks substantially increased the overall time required to preprocess the dataset, taking nearly an order of magnitude longer than the Fixed$_{800}$ configuration. Fixed-size chunks of Fixed$_{800}$, Fixed$_{1600}$, and Fixed$_{3000}$ characters all completed in under 70 seconds. Meta-chunking required significantly more processing time than any fixed-size strategy due to its complexity. Within each model, generation time was largely unaffected by the chunking strategy or the number of retrieved chunks; the dominant driver of generation time was the choice of LLM itself.
Among the tested models, Gemini-Pro was considerably slower than the others across every chunking configuration, with generation times roughly twice those of GPT-4o or Gemini Flash. This consistent gap suggests Gemini-Pro performs more computationally intensive evaluation regardless of input size.

We observed a considerably low hit rate of approximately 22\% when using smaller chunks, compared to the actual hit rate of 80\%. The hit rate represents the proportion of instances where relevant context was successfully retrieved to perform the compliance check. We could not directly compare the retrieved chunks with the ground-truth evidence documents, as it required manual checking and therefore we rather calculated the hit rate on the basis of the generated response where a No-Evidence is considered to be missing information.

\begin{table}[t]
    \centering
    \caption{Class-wise Precision, Recall, and F1-Score for compliance checking using GPT-4 model across various chunking strategies and number of chunks}
    \label{tab:classwise-gpt4}
    \begin{tabular}{lc|ccc|ccc|ccc}
        \hline
        \multirow{2}{*}{\textbf{Chunking}} & \multirow{2}{*}{\textbf{K}} & \multicolumn{3}{c|}{\textbf{Compliant}} & \multicolumn{3}{c|}{\textbf{Non-Comp.}} & \multicolumn{3}{c}{\textbf{No-Evidence}} \\        
         & & \textbf{P} & \textbf{R} & \textbf{F1} & \textbf{P} & \textbf{R} & \textbf{F1} & \textbf{P} & \textbf{R} & \textbf{F1} \\ \midrule
        \multirow{3}{*}{Fixed$_{200}$} 
         & 1 & 55 & 16 & 24 & 0 & 0 & 0 & 18 & 71 & 29 \\ 
         & 3 & 56 & 17 & 26 & 7 & 3 & 4 & 18 & 71 & 29 \\ 
         & 5 & 52 & 14 & 22 & 12 & 5 & 7 & 18 & 69 & 28 \\ \midrule
        \multirow{3}{*}{Fixed$_{800}$} 
         & 1 & 65 & 39 & 49 & 44 & 10 & 16 & 27 & 83 & 40 \\ 
         & 3 & 69 & 47 & 56 & 64 & 23 & 33 & 28 & 77 & 42 \\ 
         & 5 & 67 & 46 & 55 & 60 & 15 & 24 & 28 & 77 & 41 \\ \midrule
        \multirow{3}{*}{Fixed$_{1600}$} 
         & 1 & 63 & 31 & 41 & 58 & 18 & 27 & 24 & 83 & 38 \\ 
         & 3 & 68 & 57 & 62 & 70 & 18 & 28 & 27 & 63 & 38 \\ 
         & 5 & 70 & 62 & 66 & 82 & 23 & 35 & 30 & 66 & 41 \\ \midrule
        \multirow{3}{*}{Fixed$_{3000}$} 
         & 1 & 77 & 31 & 44 & 57 & 20 & 30 & 24 & 86 & 37 \\ 
         & 3 & 69 & 56 & 62 & 82 & 23 & 35 & 27 & 66 & 38 \\ 
         & 5 & 69 & 60 & 64 & 86 & 30 & 44 & 27 & 57 & 36 \\ \midrule
        Meta- 
         & 1 & 61 & 26 & 36 & 67 & 10 & 17 & 21 & 80 & 34 \\ 
         chunking& 3 & 65 & 41 & 50 & 89 & 20 & 33 & 26 & 80 & 40 \\ 
         & 5 & 70 & 53 & 60 & 90 & 23 & 36 & 26 & 69 & 38 \\ \midrule
    \end{tabular}
\end{table}

\begin{table}[t]
    \centering
    \caption{Class-wise Precision, Recall, and F1-Score for compliance checking using GPT-3.5 Turbo model across various chunking strategies and number of chunks}
    \label{tab:classwise-gpt35turbo}
    \begin{tabular}{lc|ccc|ccc|ccc}
        \hline
        \multirow{2}{*}{\textbf{Chunk}} & \multirow{2}{*}{\textbf{K}} & \multicolumn{3}{c|}{\textbf{Compliant}} & \multicolumn{3}{c|}{\textbf{Non-Comp.}} & \multicolumn{3}{c}{\textbf{No-Evidence}} \\        
         & & \textbf{P} & \textbf{R} & \textbf{F1} & \textbf{P} & \textbf{R} & \textbf{F1} & \textbf{P} & \textbf{R} & \textbf{F1} \\ \midrule
        \multirow{3}{*}{Fixed$_{200}$} 
         & 1 & 55 & 16 & 24 & 0 & 0 & 0 & 19 & 80 & 31 \\ 
         & 3 & 50 & 16 & 24 & 0 & 0 & 0 & 18 & 74 & 30 \\ 
         & 5 & 54 & 18 & 27 & 10 & 3 & 4 & 18 & 71 & 29 \\ \midrule
        \multirow{3}{*}{Fixed$_{800}$} 
         & 1 & 65 & 39 & 49 & 17 & 3 & 4 & 26 & 83 & 39 \\ 
         & 3 & 68 & 46 & 55 & 38 & 8 & 13 & 27 & 77 & 40 \\ 
         & 5 & 67 & 46 & 55 & 43 & 8 & 13 & 27 & 77 & 40 \\ \midrule
        \multirow{3}{*}{Fixed$_{1600}$} 
         & 1 & 61 & 45 & 52 & 50 & 3 & 5 & 27 & 77 & 40 \\ 
         & 3 & 65 & 68 & 66 & 67 & 5 & 9 & 31 & 60 & 41 \\ 
         & 5 & 62 & 68 & 65 & 100 & 5 & 10 & 27 & 49 & 35 \\ \midrule
        \multirow{3}{*}{Fixed$_{3000}$} 
         & 1 & 65 & 46 & 54 & 50 & 5 & 9 & 25 & 71 & 37 \\ 
         & 3 & 63 & 60 & 61 & 100 & 5 & 10 & 27 & 60 & 38 \\ 
         & 5 & 65 & 65 & 65 & 100 & 8 & 14 & 30 & 63 & 41 \\ \midrule        
         Meta- & 1 & 57 & 37 & 45 & 60 & 8 & 13 & 22 & 69 & 34 \\ 
         {chunking} & 3 & 59 & 47 & 53 & 80 & 10 & 18 & 23 & 60 & 33 \\ 
         & 5 & 62 & 54 & 57 & 67 & 10 & 17 & 25 & 60 & 36 \\ \midrule
    \end{tabular}
\end{table}
To investigate the optimal chunking size, Tables ~\ref{tab:classwise-gpt4}--\ref{tab:classwise-geminiPro} provide a detailed analysis of response quality across various models, utilising different chunking strategies and top-\(K\) document retrieval settings. Each table focuses on a distinct model, establishing a standardised benchmark for evaluating chunking variations. 
\begin{table}[h]
    \centering
    \caption{Class-wise Precision, Recall, and F1-Score for compliance checking using GPT-4o model across various chunking strategies and number of chunks}
    \label{tab:classwise-gpt4o}
    \begin{tabular}{lc|ccc|ccc|ccc}
        \hline
        \multirow{2}{*}{\textbf{Chunking}} & \multirow{2}{*}{\textbf{K}} & \multicolumn{3}{c|}{\textbf{Compliant}} & \multicolumn{3}{c|}{\textbf{Non-Comp.}} & \multicolumn{3}{c}{\textbf{No-Evidence}} \\        
         & & \textbf{P} & \textbf{R} & \textbf{F1} & \textbf{P} & \textbf{R} & \textbf{F1} & \textbf{P} & \textbf{R} & \textbf{F1} \\ \midrule
        \multirow{3}{*}{Fixed$_{200}$} 
         & 1 & 56 & 17 & 26 & 0 & 0 & 0 & 19 & 80 & 31 \\ 
         & 3 & 57 & 19 & 28 & 13 & 3 & 4 & 19 & 77 & 31 \\ 
         & 5 & 60 & 19 & 29 & 14 & 3 & 4 & 20 & 80 & 32 \\ \midrule
        \multirow{3}{*}{Fixed$_{800}$} 
         & 1 & 66 & 39 & 49 & 17 & 3 & 4 & 26 & 83 & 39 \\ 
         & 3 & 70 & 48 & 57 & 44 & 10 & 16 & 27 & 77 & 40 \\ 
         & 5 & 67 & 47 & 55 & 43 & 8 & 13 & 27 & 77 & 40 \\ \midrule
        \multirow{3}{*}{Fixed$_{1600}$} 
         & 1 & 66 & 44 & 53 & 40 & 5 & 9 & 27 & 83 & 41 \\ 
         & 3 & 66 & 66 & 66 & 33 & 3 & 5 & 31 & 63 & 41 \\ 
         & 5 & 67 & 72 & 69 & 33 & 3 & 5 & 35 & 63 & 45 \\ \midrule
        \multirow{3}{*}{Fixed$_{3000}$} 
         & 1 & 66 & 43 & 52 & 50 & 3 & 5 & 25 & 80 & 38 \\ 
         & 3 & 66 & 66 & 66 & 50 & 3 & 5 & 30 & 63 & 40 \\ 
         & 5 & 68 & 73 & 70 & 25 & 3 & 5 & 34 & 60 & 43 \\ \midrule
        Meta- 
         & 1 & 66 & 36 & 47 & 33 & 3 & 5 & 24 & 83 & 37 \\ 
         chunking & 3 & 68 & 48 & 57 & 50 & 3 & 5 & 28 & 83 & 41 \\ 
         & 5 & 67 & 56 & 61 & 50 & 3 & 5 & 28 & 71 & 40 \\ \midrule
    \end{tabular}
\end{table}

\begin{table}[h]
    \centering
    \caption{Class-wise Precision, Recall, and F1-Score for compliance checking using Gemini-Flash model across various chunking strategies and number of chunks}
    \label{tab:classwise-geminiFlash}
    \begin{tabular}{lc|ccc|ccc|ccc}
        \hline
        \multirow{2}{*}{\textbf{Chunking}} & \multirow{2}{*}{\textbf{K}} & \multicolumn{3}{c|}{\textbf{Compliant}} & \multicolumn{3}{c|}{\textbf{Non-Comp.}} & \multicolumn{3}{c}{\textbf{No-Evidence}} \\        
         & & \textbf{P} & \textbf{R} & \textbf{F1} & \textbf{P} & \textbf{R} & \textbf{F1} & \textbf{P} & \textbf{R} & \textbf{F1} \\ \midrule
        \multirow{3}{*}{Fixed$_{200}$} 
         & 1 & 55 & 15 & 23 & 18 & 5 & 8 & 20 & 80 & 31 \\ 
         & 3 & 58 & 17 & 26 & 18 & 5 & 8 & 20 & 80 & 32 \\ 
         & 5 & 60 & 17 & 26 & 23 & 8 & 11 & 20 & 80 & 32 \\ \midrule
        \multirow{3}{*}{Fixed$_{800}$} 
         & 1 & 70 & 41 & 51 & 17 & 3 & 4 & 25 & 80 & 38 \\ 
         & 3 & 69 & 50 & 58 & 38 & 8 & 13 & 27 & 74 & 39 \\ 
         & 5 & 67 & 48 & 56 & 58 & 18 & 27 & 28 & 74 & 41 \\ \midrule
        \multirow{3}{*}{Fixed$_{1600}$} 
         & 1 & 60 & 29 & 39 & 17 & 3 & 4 & 22 & 80 & 35 \\ 
         & 3 & 61 & 47 & 53 & 50 & 8 & 13 & 24 & 63 & 34 \\ 
         & 5 & 64 & 56 & 59 & 100 & 18 & 30 & 28 & 66 & 39 \\ \midrule
        \multirow{3}{*}{Fixed$_{3000}$} 
         & 1 & 62 & 26 & 37 & 33 & 3 & 5 & 22 & 86 & 35 \\ 
         & 3 & 64 & 49 & 56 & 50 & 5 & 9 & 24 & 66 & 35 \\ 
         & 5 & 67 & 63 & 65 & 25 & 3 & 5 & 29 & 63 & 39 \\ \midrule
        Meta-  
         & 1 & 53 & 23 & 32 & 75 & 8 & 14 & 20 & 77 & 32 \\ 
         chunking & 3 & 53 & 27 & 36 & 80 & 10 & 18 & 24 & 83 & 37 \\ 
         & 5 & 52 & 31 & 39 & 80 & 10 & 18 & 23 & 74 & 35 \\ \midrule
    \end{tabular}
\end{table}

\begin{table}
    \centering
    \caption{Class-wise Precision, Recall, and F1-Score for compliance checking using Gemini-Pro model across various chunking strategies and number of chunks}
    \label{tab:classwise-geminiPro}
    \begin{tabular}{lc|ccc|ccc|ccc}
        \hline
        \multirow{2}{*}{\textbf{Chunking}} & \multirow{2}{*}{\textbf{K}} & \multicolumn{3}{c|}{\textbf{Compliant}} & \multicolumn{3}{c|}{\textbf{Non-Comp.}} & \multicolumn{3}{c}{\textbf{No-Evidence}} \\        
         & & \textbf{P} & \textbf{R} & \textbf{F1} & \textbf{P} & \textbf{R} & \textbf{F1} & \textbf{P} & \textbf{R} & \textbf{F1} \\ \midrule
        \multirow{3}{*}{Fixed$_{200}$} 
         & 1 & 59 & 16 & 25 & 21 & 8 & 11 & 20 & 80 & 32 \\ 
         & 3 & 58 & 17 & 26 & 29 & 13 & 18 & 20 & 77 & 32 \\ 
         & 5 & 58 & 17 & 26 & 29 & 13 & 18 & 20 & 77 & 32 \\ \midrule
        \multirow{3}{*}{Fixed$_{800}$} 
         & 1 & 71 & 38 & 49 & 29 & 5 & 9 & 24 & 80 & 37 \\ 
         & 3 & 68 & 49 & 57 & 17 & 3 & 4 & 26 & 74 & 39 \\ 
         & 5 & 65 & 48 & 55 & 44 & 10 & 16 & 28 & 74 & 40 \\ \midrule
        \multirow{3}{*}{Fixed$_{1600}$} 
         & 1 & 69 & 47 & 56 & 0 & 0 & 0 & 25 & 74 & 38 \\ 
         & 3 & 66 & 46 & 54 & 25 & 5 & 8 & 26 & 74 & 39 \\ 
         & 5 & 63 & 92 & 74 & 44 & 10 & 16 & 63 & 29 & 39 \\ \midrule
        \multirow{3}{*}{Fixed$_{3000}$} 
         & 1 & 61 & 82 & 70 & 0 & 0 & 0 & 26 & 26 & 26 \\ 
         & 3 & 61 & 86 & 72 & 30 & 8 & 12 & 33 & 20 & 25 \\ 
         & 5 & 60 & 86 & 71 & 40 & 5 & 9 & 13 & 9 & 10 \\ \midrule
        Meta-  
         & 1 & 63 & 88 & 73 & 40 & 5 & 9 & 19 & 14 & 16 \\ 
         chunking & 3 & 64 & 90 & 75 & 75 & 15 & 25 & 29 & 20 & 24 \\ 
         & 5 & 63 & 83 & 72 & 56 & 23 & 32 & 33 & 23 & 27 \\ \midrule
    \end{tabular}
\end{table}

Tables~\ref{tab:classwise-gpt4}--\ref{tab:classwise-geminiPro} report class-wise precision, recall, and F1 scores for the five LLMs across the four chunking strategies and three values of K. Each table isolates one model so that chunking effects can be compared without confounding by model architecture. Across all five tables, two patterns are consistent.

First, the Fixed$_{200}$ configuration underperforms in every model. While it occasionally yields high precision on the Compliant and Non-Compliant classes, a side-effect of the model identifying very few cases and being correct on most of them, recall in these classes is uniformly low, dragging F1 to around 20--30\% across the board. Increasing K from 1 to 5 does little to help, indicating that the problem is the chunk content itself rather than the retrieval depth.

Second, increasing the chunk size has a clearly positive effect, particularly on the under-represented Non-Compliant and No-Evidence classes. The Fixed$_{1600}$ and Fixed$_{3000}$ configurations produce balanced and consistently higher F1 scores across all three classes. Meta-chunking achieves good precision on Compliant and Non-Compliant cases but suffers from low recall, driven by an inflated No-Evidence rate. The method tends to mark control questions as lacking evidence even when supporting passages are present.

A separate observation concerns models that perform poorly under direct prompting. GPT-4o and Gemini Flash miss most Non-Compliant cases without further guidance: in Table~\ref{tab:classwise-geminiFlash}, the Fixed$_{200}$ configuration shows 100\% precision on Non-Compliant but only $\sim$17\% recall, because the model correctly classifies the few cases it does flag but flags very few overall. This motivates the in-context learning experiments described next.

\begin{table*}[t]
  \caption{Improving the compliance checking capabilities of the LLMs using in-context learning}
  \label{tab:in-context}
  \centering
  \begin{tabular}{llccc|ccc|ccc|ccc}
    \toprule
    Model & Chunking & \multicolumn{3}{c|}{Compliant} & \multicolumn{3}{c|}{Non-Compliant} & \multicolumn{3}{c|}{No-Evidence} & \multicolumn{3}{c}{Weighted Average} \\
    & & P(\%) & R(\%) & F1(\%) & P(\%) & R(\%) & F1(\%) & P(\%) & R(\%) & F1(\%) & P(\%) & R(\%) & F1(\%) \\
    \midrule
    GPT-4         & Small      & 88.30 & 76.85 & 82.18 & 70.97 & 55.00 & 61.97 & 37.93 & 62.86 & 47.31 & 74.88 & 69.40 & 71.09 \\
                  & Moderate   & 81.55 & 77.78 & 79.62 & 62.86 & 55.00 & 58.67 & 37.78 & 48.57 & 42.50 & 69.09 & 67.21 & 67.94 \\
                  & Large      & 78.13 & 69.44 & 73.53 & 54.76 & 57.50 & 56.10 & 37.78 & 48.57 & 42.50 & 65.30 & 62.84 & 63.78 \\
    \midrule
    GPT-4o        & Small      & 81.52 & 69.44 & 75.00 & 43.48 & 50.00 & 46.51 & 38.64 & 48.57 & 43.04 & 65.00 & 61.20 & 62.66 \\
                  & Moderate   & 77.32 & 69.44 & 73.17 & 39.62 & 52.50 & 45.16 & 42.42 & 40.00 & 41.18 & 62.41 & 60.11 & 60.93 \\
                  & Large      & 75.00 & 63.89 & 69.00 & 38.60 & 55.00 & 45.36 & 41.18 & 40.00 & 40.58 & 60.57 & 57.38 & 58.40 \\
    \midrule
    Gemini  & Small & 76.29 & 68.52 & 72.20 & 42.86 & 52.50 & 47.19 & 41.67 & 42.86 & 42.25 & 62.36 & 60.11 & 61.00 \\
                  Flash & Moderate   & 73.27 & 68.52 & 70.81 & 36.84 & 52.50 & 43.30 & 40.00 & 28.57 & 33.33 & 58.94 & 57.37 & 57.63 \\
                  & Large      & 70.41 & 63.89 & 66.99 & 37.05 & 57.50 & 45.54 & 41.67 & 28.57 & 33.90 & 57.76 & 55.74 & 56.30 \\
    \midrule
    Gemini    & Small      & 77.78 & 77.78 & 77.78 & 61.29 & 47.50 & 53.52 & 38.64 & 48.57 & 43.04 & 66.69 & 65.57 & 65.83 \\
                  Pro & Moderate   & 73.91 & 78.70 & 76.23 & 55.56 & 50.00 & 52.63 & 37.50 & 34.29 & 35.82 & 62.94 & 63.93 & 63.35 \\
                  & Large      & 69.44 & 69.44 & 69.44 & 46.51 & 50.00 & 48.19 & 37.50 & 34.29 & 35.82 & 58.32 & 58.47 & 58.37 \\
    \bottomrule
  \end{tabular}
\end{table*}

\begin{figure*}[t]
\centering
\includegraphics[width=1\linewidth]{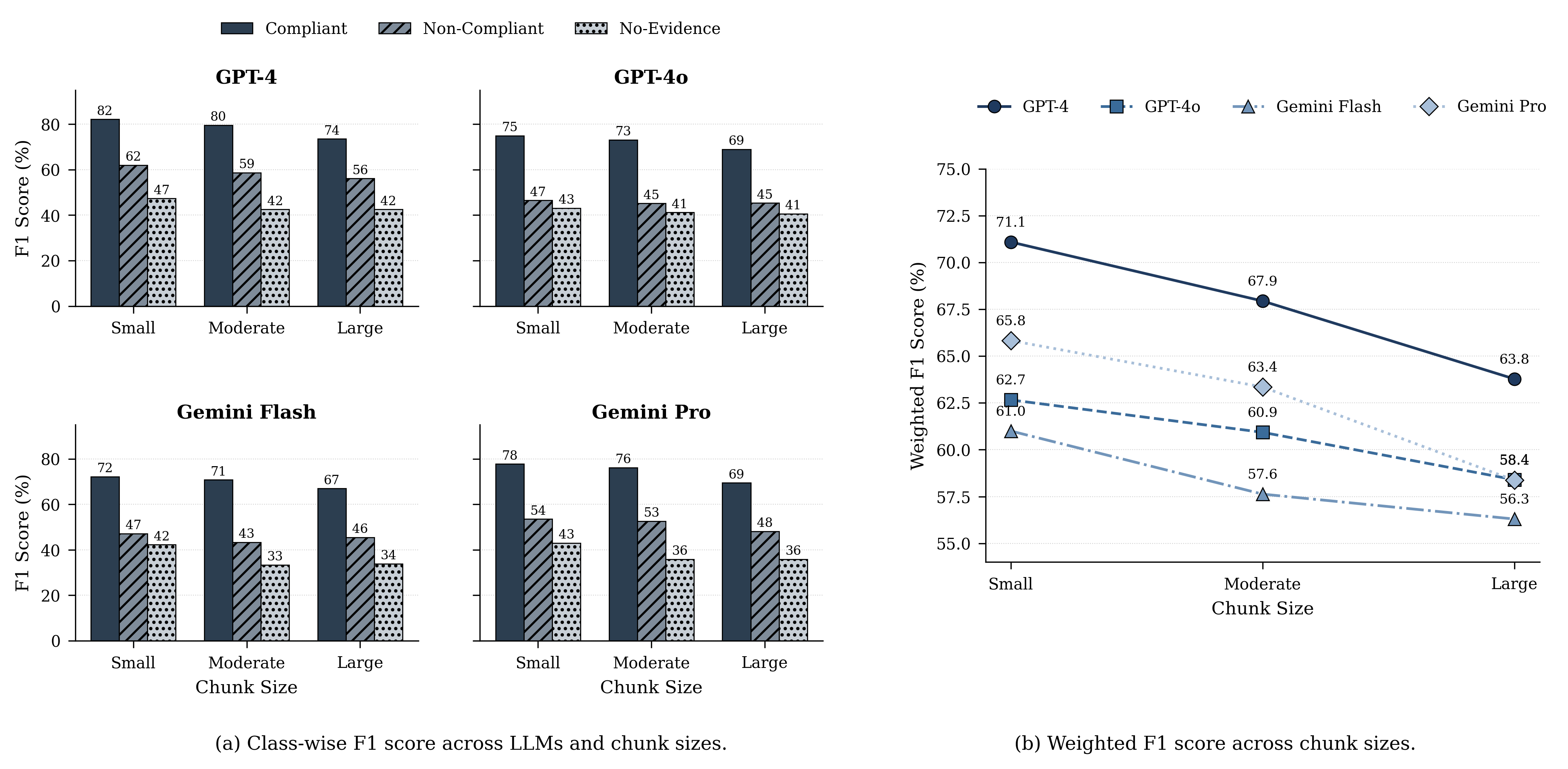}
\caption{F1 score performance after in-context learning across four LLMs and three chunk sizes.}
\label{fig:tab8}
\end{figure*}

In our experiments, we also explored the use of in-context learning to enhance the decision-making capabilities of a model and improve the overall response quality. Analysing error cases from the direct prompting approach revealed consistent shortcomings in the ability of a model to determine control compliance accurately based on the retrieved document information. Although relevant information was often present, the model struggled to align its decisions with human annotators, particularly in nuanced cases such as indirect compliance.
One such scenario involved vendors leveraging third-party services, such as cloud service providers, that complied with specific controls. Human annotators considered such cases compliant because the third-party compliance indirectly satisfied the control requirements. However, our CTRAG framework classified these cases as non-compliant, as the vendor itself did not directly adhere to the control requirements. By incorporating in-context learning, we were able to guide the model in adapting to these scenarios, enabling it to mimic human-like decision-making more effectively.

In-context learning capabilities were also employed to address areas where the model consistently made errors. By providing detailed contextual examples that illustrated specific decision-making rules, the ability of a model to classify cases improved significantly. 
Table \ref{tab:in-context} and Figure~\ref{fig:tab8} demonstrate this improvement, highlighting the importance of in-context learning in aligning the model responses with human judgment. As Figure~\ref{fig:tab8} (a) shows, every model exhibits the same class ranking; Compliant scores highest, No-Evidence lowest, regardless of chunk size, suggesting that the difficulty hierarchy is intrinsic to the task rather than a quirk of any single model or chunking strategy. Figure~\ref{fig:tab8} (b) further shows that weighted F1 declines monotonically as chunk size grows across all four models, with small chunks consistently outperforming larger ones.

The configurations reported in Table \ref{tab:in-context} reflect a deliberate narrowing of the experimental space. We omitted the 200-character chunk size, as its performance was consistently low across all metrics in the direct-prompting experiments, and increasing the retrieval depth failed to improve its effectiveness. Meta-chunking was also excluded due to its prohibitive chunking time, which made it unsuitable for this particular use case.

For these experiments, we fixed the retrieval depth at $K=4$ based on the findings from the direct-prompting experiments. Across the evaluated models, retrieval depths between $K=3$ and $K=5$ consistently provided the best balance between contextual completeness and retrieval noise. Selecting $K=4$ allowed us to capture sufficient supporting evidence for compliance assessment while limiting the inclusion of irrelevant information that could negatively affect generation quality. This setting therefore represents a practical trade-off between retrieval effectiveness and computational efficiency.
Using this configuration, Fixed$_{800}$ chunks yielded the strongest overall results across the compliance classes, particularly when paired with GPT-4. Similar trends were observed across the other models, supporting our objective of maintaining smaller chunk sizes to minimise token consumption while preserving retrieval effectiveness.

\begin{table}
  \caption{Performance Comparison Across Models for Overall Metrics}
  \label{tab:overall-performance-comparison}
  \centering
  \begin{tabular}{llccc}
    \toprule
    Model          & Chunking & Precision (\%) & Recall (\%) & F1 (\%) \\
    \midrule
    GPT-4          & Small             & 71.91          & 85.33       & 78.05   \\
                   & Moderate          & 70.00          & 74.67       & 72.26   \\
                   & Large             & 62.07          & 72.00       & 66.67   \\
    \midrule
    GPT-4o         & Small             & 63.74          & 77.33       & 69.88   \\
                   & Moderate          & 61.63          & 70.67       & 65.84   \\
                   & Large             & 57.14          & 69.33       & 62.65   \\
    \midrule
    Gemini Flash   & Small             & 60.47          & 69.33       & 64.60   \\
                   & Moderate          & 58.54          & 64.00       & 61.15   \\
                   & Large             & 54.12          & 61.33       & 57.50   \\
    \midrule
    Gemini Pro     & Small             & 68.00          & 68.00       & 68.00   \\
                   & Moderate          & 66.18          & 60.00       & 62.94   \\
                   & Large             & 56.00          & 56.00       & 56.00   \\
    \bottomrule
  \end{tabular}
\end{table}

A significant challenge was the low recall for the Non-Compliant class, which indicates the risk of missing some compliance fail cases, a critical concern for compliance checking. Many of these missed cases were instead classified under the No-Evidence class, reflecting situations where the model failed to find relevant information to determine compliance. In the final solution, however, No-Evidence classifications were reclassified as Non-Compliant, as the absence of relevant information typically implies that the vendor does not comply with the rules and regulations and therefore this information is not available in the documents. This reclassification approach significantly improved recall for the Non-Compliant class avoiding expensive fine-tuning for the task, as shown in Table \ref{tab:overall-performance-comparison}. By addressing the No-Evidence cases in this manner, we captured many previously missed compliance fail cases, achieving a more robust and realistic compliance-checking solution. The final results presented in Table \ref{tab:overall-performance-comparison} demonstrate an acceptable level of compliance-checking performance, aligning well with our objectives and the task requirements. 

\begin{figure}[h]
  \centering
  \includegraphics[width=1\linewidth]{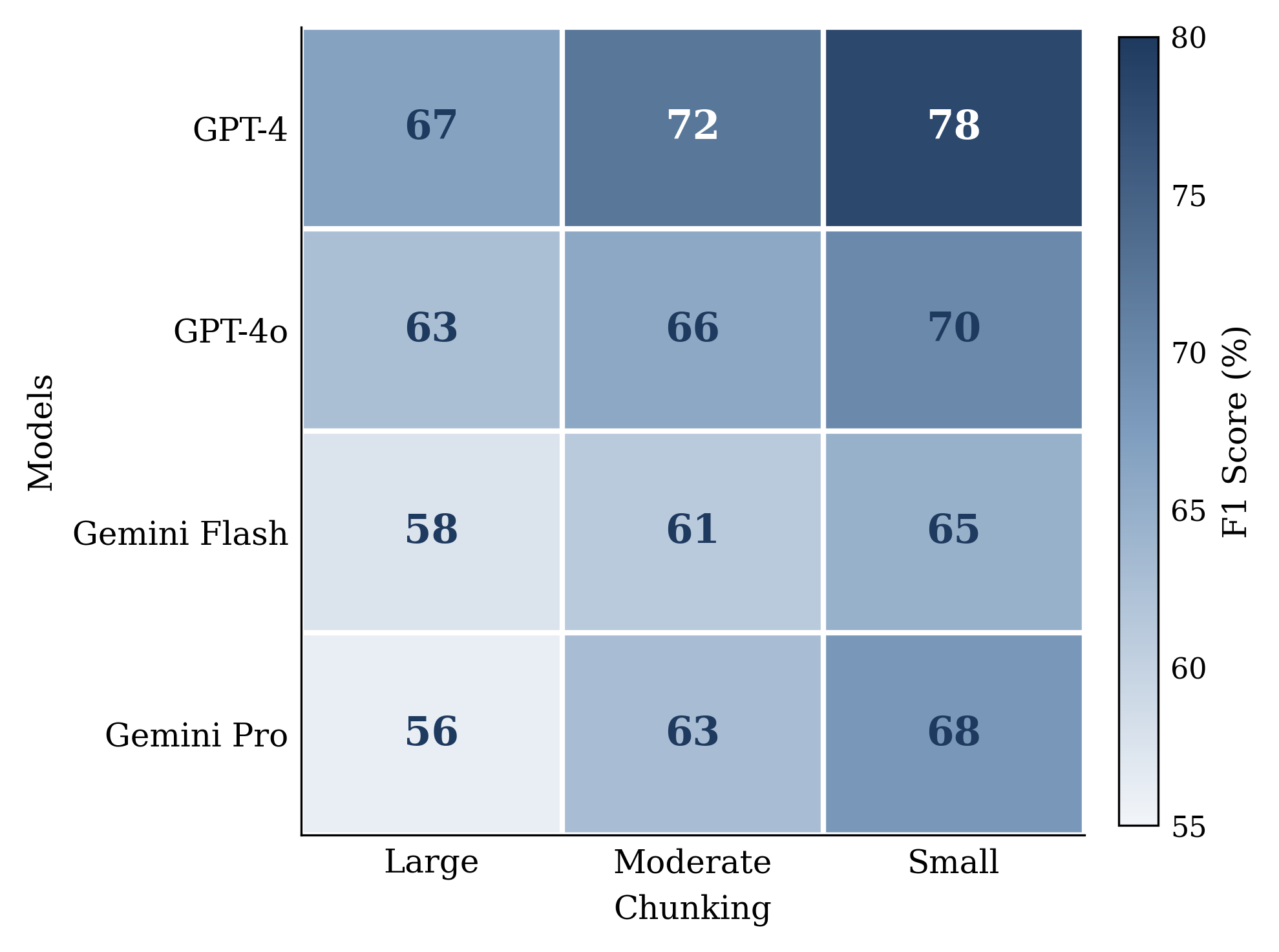}
  \caption{Heatmap of F1 scores across models and chunking strategies, highlighting performance variations}
  \label{fig:heatmap}
\end{figure}

The heatmap in Figure \ref{fig:heatmap} highlights that small chunks with \(K=4\) consistently deliver better performance across all models, emphasising that increasing the chunk size does not necessarily enhance compliance-checking capabilities. This finding may be attributed to two significant factors.
First, the "needle-in-a-haystack" effect becomes more prominent with larger chunks. When the model is presented with a substantial amount of context, the relevant information might get buried within a sea of less pertinent details. This overwhelming context can make it challenging for the model to extract and synthesize the key information required to arrive at a conclusive decision, thereby reducing its effectiveness.
Second, larger chunks inherently contain more irrelevant or noisy text. This additional noise increases the complexity of the retrieval task, as the retriever might struggle to focus on the most relevant portions of the text. Consequently, this can lead to situations where the retriever misses critical chunks that contain the necessary evidence for compliance checking. The combination of increased noise and the difficulty of a model in narrowing down essential information could explain the diminished performance observed with larger chunks.
These observations underscore the importance of selecting an optimal chunk size that balances contextual richness with retrievability and processing efficiency, ensuring the ability to focus on the most relevant information while minimising noise.

\section{Conclusion}
\label{section:conclusion}

This paper investigated the impact of chunking strategies, retrieval configurations, LLM architectures, and in-context learning on automated compliance checking. Across the evaluated configurations, small chunks with $K=4$ achieved the most consistent performance, while larger chunks generally suffered from reduced precision and recall due to context dilution. Meta-chunking, while intended to preserve logical coherence, incurred prohibitive preprocessing costs and was not competitive with fixed-size chunking on this task. These findings suggest that retrieval quality and evidence granularity are more important than increasing context size for compliance-oriented document analysis. In-context learning substantially improved how the models handled cases that direct prompting struggled with, particularly indirect compliance via third-party services. Gains were most pronounced for the under-represented Non-Compliant and No-Evidence classes, where targeted examples helped the model adopt the decision rules used by human annotators.
The reported deployment in a Big Four professional services firm produced an approximately 60\% reduction in manual reviewer effort relative to a firm existing process.
These findings suggest the approach is well suited to scaling beyond the single-organization deployment evaluated here, and to adjacent document-analysis tasks such as policy gap analysis and contract review.
The study has several limitations. The evaluation is based on 240 controls and 45 documents from a single organization; broader testing across regulatory domains is needed to establish how well the chunking and ICL findings generalize. The dataset is proprietary and cannot be released. Inter-annotator agreement on the binary Pass/Fail labels was not measured, and the final pipeline relies on a conservative No-Evidence-to-Non-Compliant reclassification that trades precision for recall. Future work will address these limitations, run statistical significance tests against the chunking and K configurations, and extend CTRAG with proper Meta-Chunking variants, semantic chunking and non-RAG baselines.

\section{Acknowledgments}
\label{section:ack}
This research is supported by the Advanced Research and Engineering Centre (ARC) in Northern Ireland, funded by PwC and Invest NI. The views expressed are those of the authors and do not necessarily represent those of ARC or the funding organisations.

The authors appreciate the use of the Kelvin2 High Performance Computing cluster at Queen's University Belfast, and the cloud services and APIs provided by PwC for computational work.


\bibliography{references}

\section*{Appendix}
\appendix
\subsection{Complete K-level breakdown of chunking and generation time}
\label{app1}

\begin{table}[h]
  \caption{Chunking and Generation Times Across Models and Chunking Strategies}
  \label{tab:table_time}
  \centering
  \begin{tabular}{l|c|ccc|c}
    \toprule
    \textbf{Chunking} & \textbf{Chunking} & \multicolumn{3}{c|}{\textbf{Generation Times}} & \textbf{Avg.} 
    \\
    \textbf{Strategy} & \textbf{Time} & K=1 & K=3 & K=5 & \\
    \midrule
    \multicolumn{6}{c}{\textbf{GPT-3.5 Turbo}} \\ \midrule
    Fixed$_{200}$ & 263 & 98 & 100 & 103 & 100 \\
    Fixed$_{800}$ & 65 & 101 & 102 & 103 & 102 \\
    Fixed$_{1600}$ & 42 & 108 & 110 & 111 & 110 \\
    Fixed$_{3000}$ & 27 & 109 & 110 & 113 & 111 \\
    Meta-chunking & 1674 & 105 & 106 & 108 & 106 \\
    \midrule
	\multicolumn{6}{c}{\textbf{GPT-4}} \\ \midrule
    Fixed$_{200}$ & 263 & 107 & 109 & 111 & 110 \\
    Fixed$_{800}$ & 65 & 105 & 106 & 108 & 107 \\
    Fixed$_{1600}$ & 42 & 114 & 115 & 117 & 115 \\
    Fixed$_{3000}$ & 27 & 119 & 121 & 122 & 121 \\
    Meta-chunking & 1674 & 117 & 118 & 120 & 119 \\
    \midrule
	\multicolumn{6}{c}{\textbf{GPT-4o}} \\ \midrule
    Fixed$_{200}$ & 263 & 92 & 93 & 95 & 94 \\
    Fixed$_{800}$ & 65 & 81 & 82 & 84 & 82 \\
    Fixed$_{1600}$ & 42 & 82 & 84 & 85 & 84 \\
    Fixed$_{3000}$ & 27 & 84 & 85 & 87 & 85 \\
    Meta-chunking & 1674 & 80 & 81 & 83 & 82 \\
    \midrule
	\multicolumn{6}{c}{\textbf{Gemini Flash}} \\ \midrule
    Fixed$_{200}$ & 263 & 88 & 90 & 92 & 90 \\
    Fixed$_{800}$ & 65 & 75 & 76 & 78 & 76 \\
    Fixed$_{1600}$ & 42 & 76 & 77 & 79 & 77 \\
    Fixed$_{3000}$ & 27 & 80 & 81 & 83 & 82 \\
    Meta-chunking & 1674 & 79 & 81 & 82 & 81 \\
    \midrule
	\multicolumn{6}{c}{\textbf{Gemini Pro}} \\ \midrule
    Fixed$_{200}$ & 263 & 198 & 200 & 203 & 200 \\
    Fixed$_{800}$ & 65 & 171 & 172 & 174 & 173 \\
    Fixed$_{1600}$ & 42 & 167 & 169 & 171 & 169 \\
    Fixed$_{3000}$ & 27 & 156 & 158 & 160 & 158 \\
    Meta-chunking & 1674 & 191 & 193 & 196 & 194 \\
    \bottomrule
  \end{tabular}
\end{table}
\end{document}